# Nonparametric Nearest Neighbor Descent Clustering based on Delaunay Triangulation

Teng Qiu     (qiutengcool@163.com)
Yongjie Li*     (liyj@uestc.edu.cn)

Key Laboratory of NeuroInformation, Ministry of Education of China, School of Life Science and Technology, University of Electronic Science and Technology of China, Chengdu, China
*Corresponding author.

**Abstract**: In our physically inspired in-tree (IT) based clustering algorithm and the series after it, there is only one free parameter involved in computing the potential value of each point. In this work, based on the Delaunay Triangulation or its dual Voronoi tessellation, we propose a nonparametric process to compute potential values by the local information. **This computation, though nonparametric, is relatively very rough, and consequently, many local extreme points will be generated. However, unlike those gradient-based methods, our IT-based methods are generally insensitive to those local extremes. This positively demonstrates the superiority of these parametric (previous) and nonparametric (in this work) IT-based methods**.

## 1 Introduction

In (*1*), we proposed a physically inspired clustering algorithm, in which an in-tree (IT) structure was first constructed. This IT structure organizes the data points into the clusters with several undesired connections (edges) between them requiring to be removed. In (*1*) and the series (*2-4*) after it, we proposed several methods to remove those redundant edges, either by using semi-supervised and interactive strategies or by combining with the Decision Graph recently proposed by Rodriguez and Laio (5) and some other popular methods as affinity propagation (AP) (6) and Isomap (7). All these efforts seem quite effective, showing the strong extensibility of this IT structure. For example, in (3), the proposed method (G-AP) lets AP do the post processing to the IT structure, which overcomes the weakness of AP in non-spherical cluster detection. In (4), the proposed method (IT-map) replaces the K-nearest-neighbor structure in Isomap by our IT structure and maps the IT structure into a low-dimensional space, while guaranteeing clusters to be distinguishable, which is generally hard for Isomap due to the so-called crowding problem (8).

## 2 Motivation

However, both our IT-based methods and Rodriguez and Laio's method involve one free parameter in computing either potential variable (ours) or density variable (Rodriguez and Laio). Quantitatively, the potential values are inversely proportional to density values, so the solutions to nonparameterization should work for both.

How to compute the potential or density of data points with no free parameter

involved[1]? The Delaunay Triangulation (9) used in our another recent paper (10) may shed light on it, considering the fact (as in Fig. 1A) that the volumes of the basic lattices (or $d$-simplex) is visually quite consistent with the density distribution of the dataset. In fact, the Delaunay Triangulation and its dual Voronoi Tessellation (Fig. 1B) have wide applications in density estimation (11, 12). This builds the basis for us to fulfill our idea.

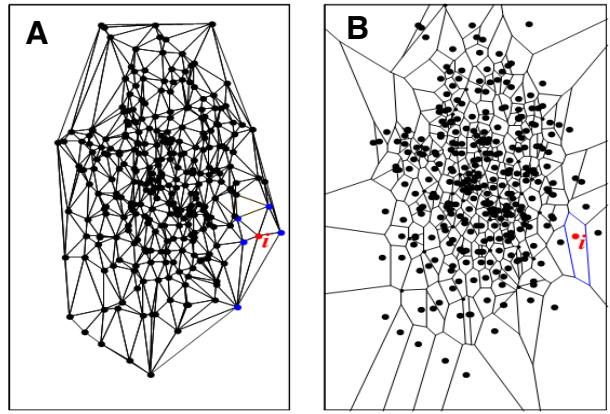

**Fig. 1. The Delaunay (A) and Voronoi (B) tessellation ($d = 2$) for the same dataset.** (A) This Delaunay graph consists of triangulars. The neighbors of node $i$ are denoted in blue. (B) Each point is inside a basic lattice, called Voronoi cell (area enclosed by the blue edges).

### 3 Method

**Step 1, compute the potentials of all data points.** For a dataset in $d$-dimensional Euclidean space, we first construct its Delaunay Triangulation (DT), or its dual Voronoi Tessellation (VT). The potential of each point is then defined as

$$P_i = f(S_i), \quad (1)$$

where the function $f(x) = x$, or other monotonous functions[2]. As the variants[3] of the density defined in (12, 13), $S_i$ can be set as (**i**) the total volume of the neighboring lattices of node $i$ in DT, or (**ii**) the volume of Voronoi cell of node $i$ in VT. In fact, we can go a step further to define $S_i$ as (**iii**) the median (or other forms[4]) of the distances between node $i$ with all its neighbor nodes in DT, which is much faster than the above two definitions, especially when dealing with high-dimensional dataset. See an illustration ($d = 2$) for these Four methods in computing $S_i$ in Fig. 1.

**Step 2, the nearest neighbor descent.** For any node $i$, its directed node is defined as

$$I_i = \min \underset{j \in J_i}{\arg\min} \, D_{i,j},$$

---

[1] Such methods without any parameter may not necessarily be superior to the ones with one parameter in all cases.
[2] Other monotonous-increasing functions as $-e^{-x}$, $\log(1+x)$ and $1/(1+e^{-x})$ will lead to the same result (i.e., IT structure) as f(x)=x, since they don't change the relative relationships of potential values and thus hardly make any effect on step 2. The main role of them is to increase the visual difference (if needed) of potential values on nodes, as in Fig. 2.
[3] Density is defined as the *inverse* to the total volume of the neighboring lattices of node $i$ in DT (In Schaap's paper), or as the inverse to the volume of Voronoi cell (in Ramella's paper).
[4] In practice, other statistics as mean, max, min, sum can also be appropriate, to some degree.

where [5] $J_i = \{j \mid P_j < P_i\} \cup \{j \mid P_j = P_i \,\&\&\, j < i\}$, and $D_{i,j}$ is the distance (e.g., Euclidean distance) between nodes *i* and *j*. If $J_i$ is null, then set $I_i = i$. Consequently, an IT structure is obtained.

**Step 3, cut the undesired (or redundant) edges between clusters.** In general, any cutting method in our previous works (1-4) can be used here.

**Step 4, identify the root node of each node.** Step 3 divides the IT structure into separate sub-ITs, each representing one cluster. Each sub-IT has one root node. Each root node can be viewed as the representative (or exemplar) of each cluster. Since any other node has one and only one directed path to reach the root node in the same sub-IT, the root of each node can be identified by searching along the directed path and consequently the nodes with the same root node are assigned to the same cluster.

## 4 Experiments

In Fig. 2, we show the Delaunay Triangulations for several 2D datasets (5, 14, 15) in the upper panels and the corresponding IT structures in the lower panels. In Figs. 3 and 4, we respectively show the performance of two cutting methods:

**Nonparametric-IT-SS-II[6]:** Figure 3 shows the results of the semi-supervised (SS) cutting method based on Cell-like divisive mechanism (2). This divisive mechanism takes the labeled data as "genetic material" and each independent IT structure as one cell. The process of edge cutting can thus be viewed as that of cell dividing. Although the edges are explored according to their lengths (in decreasing order), whether one edge is to be removed or not is depended on whether this cutting behavior contributes to dividing the "impure cells" (containing more than one kind of genetic material) into purer[7] ones. See details in (2). Consequently, only the edges between clusters are removed, as desired. All the obtained clusters contain only one kind of labeled data points. The pleasing results (Fig. 3, lower panels) demonstrate that this semi-supervised cutting method is quite suitable for the proposed nonparametric strategy.

**Nonparametric-IT-DG:** Figure 4 shows the results of the interactive cutting method based on Rodriguez and Laio's Decision Graph (DG). However, we show (Fig. 4, upper panels) the Decision graphs based on the variables, Potential (P) and edge length (W), from our IT structures (Fig. 2, lower panels). We only need to select out the pop-out points with low values of variable P and high values of variable W. The directed edges in IT that are started from the corresponding nodes of the identified points will be removed. The eventual clustering results (Fig. 4, lower panels) are all quite consistent with visual perception, except the last one.

Note that the tests in Figs. 2~4 are based on the 1st definition of $S_i$. Comparable results when $S_i$ takes the 3rd definition are shown in Fig. S1.

---

[5] Note that we don't restrict node *i* to select its the directed node only among its neighbors in the graph obtained in step 1 (a lesson learned from our previous work). Moreover, the indexes of data points are used as a complementary of potential variable, mainly functioning to the close points, especially the overlapped ones, with the same potential, for which indexes can make only one of them as a representative to build up connection with other nodes, so as to guarantee the redundant connection (i.e., edge) between clusters generally no more than one.

[6] We denote the semi-supervised cutting method in ref.1 as type I.

[7] The cell without any genetic material is not allowed.

**Discussions**

**Problem:** for our previous cutting methods as IT-SS, IT-DG, IT-map, IT-AP (or G-AP), (i) there is at least one type of data set (*8*) as in Fig. S2, that is hard for any of them to cluster successfully in this nonparametric framework; (ii) not all of them are well suitable for this kind of nonparametric proposal. The best suitable one should be IT-SS. Next are IT-DG and IT-map. However, for G-AP, the performances are generally much worse than that in the previous kernel-based solutions.

**Comparing with gradient-based solutions**: the gradient-based solutions are routinely used by researchers directly or indirectly in density-based (16, 17) (18), graph-based (19), and physics-based (20) methods in which cluster centers are equated with the extreme points or density peaks irrespective of they are local ones or not. Therefore, besides their own problems these methods have one common problem of being very sensitive to the estimation of the density or its inverse form—the potential. Since the potentials in this work are so roughly estimated in Eq. 1, which should lead to many local and fake extreme points in terms of potential variable. For those gradient-based methods, this means many fake clusters would occur. In comparison, the proposed method abandons the gradient-based solution, yet generally avoiding that problem as shown in Figs. 2~5. Therefore, this work not only provides an optional way for us to avoid the free parameter in our previous methods. ***More importantly, it also demonstrates in a positive way the superiority of the strategy (i.e. the nearest neighbor descent) in (1) for constructing the IT structure.***

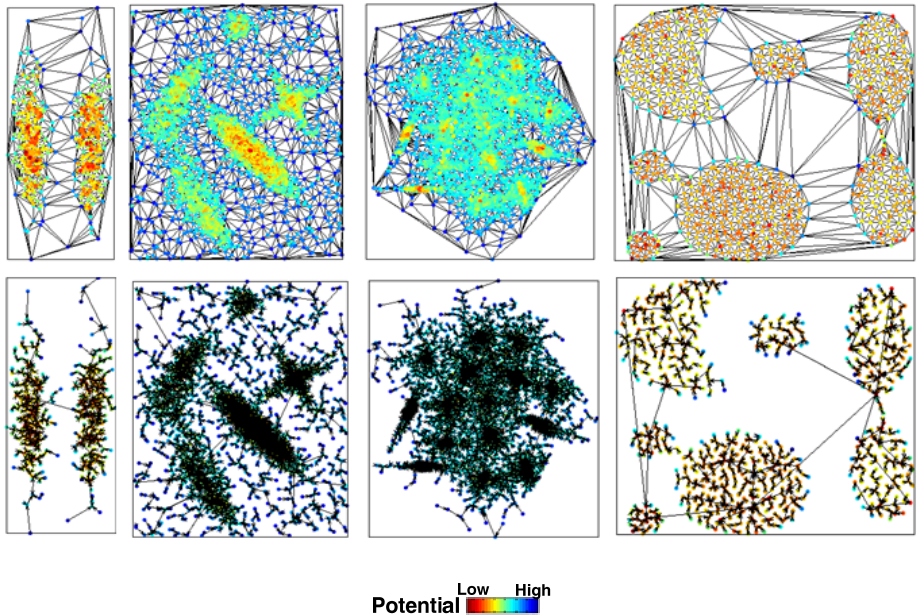

**Fig. 2. The Delaunay Graphs (upper panels) and the corresponding IT structures (lower panels) for four 2D datasets ($S_i$ takes the 1st definition).** For all images, different colors on nodes denote different potential values on nodes. Note that all IT

structures in lower panels are not the sub-graphs of the corresponding Delaunay Graph in upper panels, since the connections in upper panels are not used in step 2. For all datasets, we choose $f(S_i) = log(1+ S_i/\min_i S_i)$ in Eq. 1 to compute potential values.

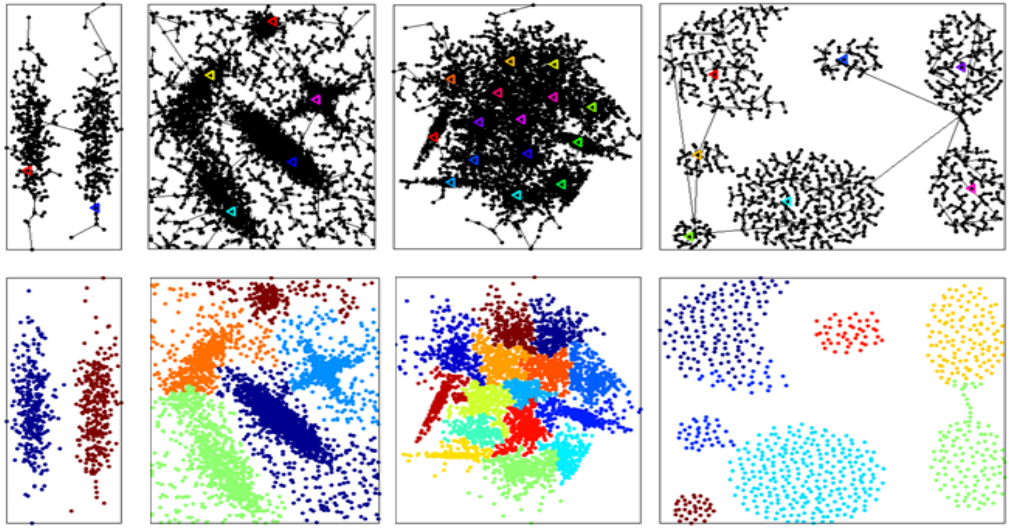

**Fig. 3. Results of nonparametric-IT-SS-II. Upper panels:** triangulars denote the labeled points. Each cluster is assumed to have one point labeled. Labels from different clusters are differentiated by different colors. Supervised by these very few labeled data points, the connections between clusters will be automatically removed. **Lower panels:** the corresponding clustering results.

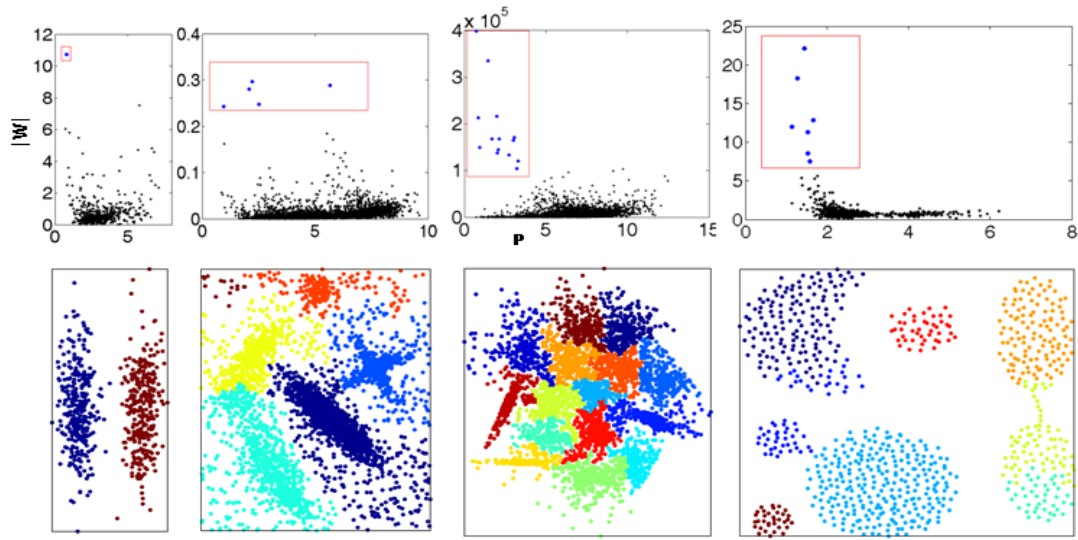

**Fig. 4. Results of nonparametric-IT-DG. Upper panels**: the IT-based Decision Graphs. The pop-out points (blue) correspond to the identified start nodes of the redundant edges in IT structure. Note that the number of the pop-out points in each Decision graph is one less than that in Rodriguez and Laio's Decision Graph. **Lower panels:** the corresponding clustering results.

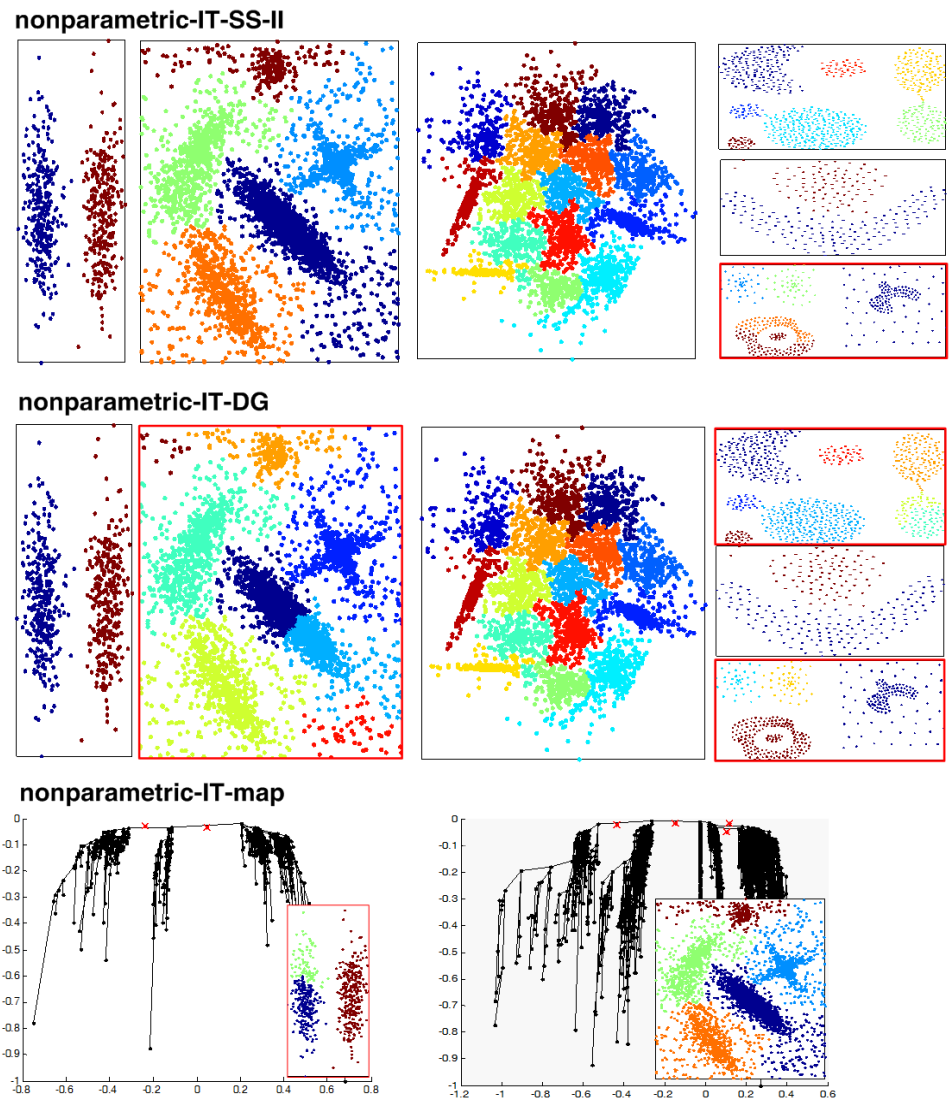

**Fig. S1. Results[8] for different cutting methods ($S_i$ takes the 3rd definition). Panels in red box denote the problematic results. We only test the first two datasets for IT-map. It shows that the performance of the embeddings are worse than that in ref. 4.**

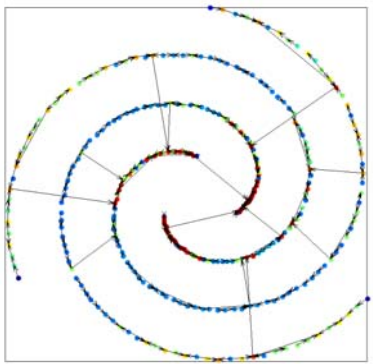

**Fig. S2. The IT structure of a spiral dataset ($S_i$ takes the 3rd definition).** Colors on nodes denote different potential values. This is hard for any previous cutting method to obtain a satisfactory result. However, if the number of labeled data points is large enough, IT-SS-II can still obtain an ideal result.

---

[8] All these datasets are downloaded from http://cs.joensuu.fi/sipu/datasets/

# References

1. Qiu T, Yang K, Li C, & Li Y (2014) A Physically Inspired Clustering Algorithm: to Evolve Like Particles. *arXiv preprint arXiv:1412.5902*.
2. Qiu T & Li Y (2014) An Effective Semi-supervised Divisive Clustering Algorithm. *arXiv preprint arXiv:1412.7625*.
3. Qiu T & Li Y (2015) A Generalized Affinity Propagation Clustering Algorithm for Nonspherical Cluster Discovery. *arXiv preprint arXiv:1501.04318*.
4. Qiu T & Li Y (2015) IT-map: an Effective Nonlinear Dimensionality Reduction Method for Interactive Clustering. *arXiv preprint arXiv:1501.06450*.
5. Rodriguez A & Laio A (2014) Clustering by fast search and find of density peaks. *Science* 344(6191):1492-1496.
6. Frey BJ & Dueck D (2007) Clustering by passing messages between data points. *Science* 315(5814):972-976.
7. Tenenbaum JB, De Silva V, & Langford JC (2000) A global geometric framework for nonlinear dimensionality reduction. *Science* 290(5500):2319-2323.
8. Van der Maaten L & Hinton G (2008) Visualizing data using t-SNE. *Journal of Machine Learning Research* 9(2579-2605):85.
9. Delaunay B (1934) Sur la sphere vide. *Izv. Akad. Nauk SSSR, Otdelenie Matematicheskii i Estestvennyka Nauk* 7(793-800):1-2.
10. Qiu T & Li Y (2015) Clustering by Descending to the Nearest Neighbor in the Delaunay Graph Space. *arXiv preprint arXiv:1502.04502*.
11. Barr CD & Schoenberg FP (2010) On the Voronoi estimator for the intensity of an inhomogeneous planar Poisson process. *Biometrika* 97(4):977-984.
12. Schaap WE (2007) The Delaunay tessellation field estimator. (Ph. D. thesis, Groningen University).
13. Ramella M, Nonino M, Boschin W, & Fadda D (1998) Cluster identification via voronoi tessellation. *arXiv preprint astro-ph/9810124*.
14. Fränti P & Virmajoki O (2006) Iterative shrinking method for clustering problems. *Pattern Recognit.* 39(5):761-775.
15. Gionis A, Mannila H, & Tsaparas P (2007) Clustering aggregation. *ACM Trans. Knowl. Discovery Data* 1(1):4.
16. Fukunaga K & Hostetler L (1975) The estimation of the gradient of a density function, with applications in pattern recognition. *IEEE Trans. Inf. Theory* 21(1):32-40.
17. Cheng Y (1995) Mean shift, mode seeking, and clustering. *IEEE Trans. Pattern Anal. Mach. Intell.* 17(8):790-799.
18. Hinneburg A & Keim DA (1998) An efficient approach to clustering in large multimedia databases with noise. *Proceedings of the 4th International Conference on Knowledge Discovery and Data Mining, R. Agrawal, P.E. Stolorz, G. Piatetsky-Shapiro, Eds.*, pp 58-65.
19. Koontz WL, Narendra PM, & Fukunaga K (1976) A graph-theoretic approach to nonparametric cluster analysis. *IEEE Trans. Comput.* 100(9):936-944.
20. Ruta D & Gabrys B (2009) A framework for machine learning based on dynamic physical fields. *Natural Computing* 8(2):219-237.